\documentclass[conference]{IEEEtran}
\usepackage{times}

\usepackage[numbers]{natbib}
\usepackage{multicol}
\usepackage[bookmarks=true]{hyperref}

\usepackage[usenames,dvipsnames]{xcolor}
\usepackage[english]{babel}
\usepackage{blindtext}

\usepackage{tikz} % pm command
\usepackage{bm}
\usepackage{float}
\usepackage{siunitx}

\hypersetup{
    colorlinks,
    linkcolor={black},
    citecolor={black},
    urlcolor={black}
}

\usepackage{systeme}
\usepackage{amsmath, cases}
\usepackage{graphicx}
\usepackage{esvect}

\usepackage[normalem]{ulem}

\begin{document}

% paper title
\title{Beyond Flat GelSight Sensors: Simulation of Optical Tactile Sensors of Complex Morphologies for Sim2Real Learning}

% \url{https://danfergo.github.io/geltip-sim}

% You will get a Paper-ID when submitting a pdf file to the conference system
%\author{Author Names Omitted for Anonymous Review. Paper-ID [205]}

\author{

\authorblockN{Daniel Fernandes Gomes}
\authorblockA{Department of Computer Science, University of Liverpool\\Department of Engineering, King's College London\\
danfergo@kcl.ac.uk}
\and
\authorblockN{Paolo Paoletti}
\authorblockA{School of Engineering,\\ University of Liverpool\\
paolo.paoletti@liverpool.ac.uk}
\and
\authorblockN{Shan Luo}
\authorblockA{Department of Engineering,\\ King's College London\\
shan.luo@kcl.ac.uk}}

% avoiding spaces at the end of the author lines is not a problem with
% conference papers because we don't use \thanks or \IEEEmembership

% for over three affiliations, or if they all won't fit within the width
% of the page, use this alternative format:
% 
%\author{\authorblockN{Michael Shell\authorrefmark{1},
%Homer Simpson\authorrefmark{2},
%James Kirk\authorrefmark{3}, 
%Montgomery Scott\authorrefmark{3} and
%Eldon Tyrell\authorrefmark{4}}
%\authorblockA{\authorrefmark{1}School of Electrical and Computer Engineering\\
%Georgia Institute of Technology,
%Atlanta, Georgia 30332--0250\\ Email: mshell@ece.gatech.edu}
%\authorblockA{\authorrefmark{2}Twentieth Century Fox, Springfield, USA\\
%Email: homer@thesimpsons.com}
%\authorblockA{\authorrefmark{3}Starfleet Academy, San Francisco, California 96678-2391\\
%Telephone: (800) 555--1212, Fax: (888) 555--1212}
%\authorblockA{\authorrefmark{4}Tyrell Inc., 123 Replicant Street, Los Angeles, California 90210--4321}}

\maketitle

\begin{abstract}
Recently, several morphologies, each with its advantages, have been proposed for the \textit{GelSight} high-resolution tactile sensors. However, existing simulation methods are limited to flat-surface sensors, which prevents its usage with the newer sensors of non-flat morphologies in Sim2Real experiments.
In this paper, we extend a previously proposed GelSight simulation method, which was developed for flat-surface sensors, and propose a novel method for curved sensors. In particular, we address the simulation of light rays travelling through a curved tactile membrane in the form of geodesic paths. The method is validated by simulating the finger-shaped GelTip sensor and comparing the generated synthetic tactile images against the corresponding real images. Our extensive experiments show that combining the illumination generated from the geodesic paths, with a background image from the real sensor, produces the best results when compared to the lighting generated by direct linear paths in the same conditions. As the method is parameterised by the sensor mesh, it can be applied in principle to simulate a tactile sensor of any morphology. 
The proposed method not only unlocks simulating existing optical tactile sensors of complex morphologies, but also enables experimenting with sensors of novel morphologies, before the fabrication of the real sensor. \\
Project website: \url{https://danfergo.github.io/geltip-sim}\\
\end{abstract}

\IEEEpeerreviewmaketitle

\section{Introduction}

Tactile sensing is an important capability for robots that interact with objects in unstructured environments. In such challenging environments, vision often produces imprecise or incomplete observations, due to occlusion of the objects by the robot hands, changes of light conditions and object colours variances, which makes tactile sensing vital in determining object properties. However, the design of optimal tactile sensors still remains an open challenge and there are open questions for their morphology, the fitting of the necessary electronics in compact volumes, and the required soft membrane resistant to wear and tear, etc. To address these questions, in the past decades a wide variety of working principles and designs have been explored~\citet{dahiya2009tactile,SurveyTactileSensingShan,li2020review}. 

\begin{figure}[t]
\centering
\includegraphics[width=0.49\textwidth]{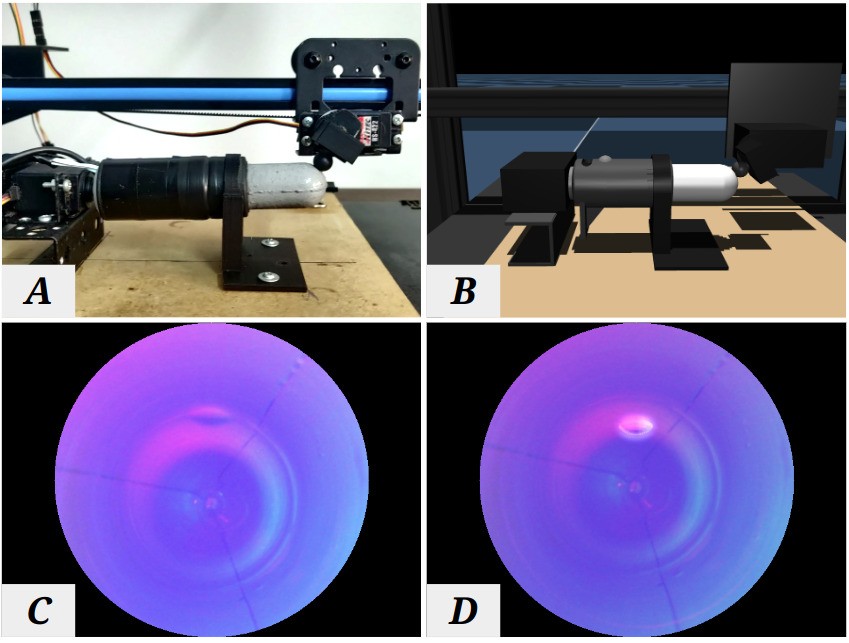}
\caption{\textbf{(A)} and \textbf{(B)}: The real and the simulated experimental setups respectively, in each a GelTip tactile sensor is mounted onto a 3D printer and contacts a 3D printed object (here is a cube with a hollow cylinder in the centre). \textbf{(C)} and \textbf{(D)}: The corresponding tactile images captured using the GelTip sensor, real and simulated (with \textit{Plane} light field) respectively.}
\label{fig:cover}
\end{figure}

One convenient and promising direction is the usage of standard cameras behind a soft opaque membrane to capture the up-close and colour-invariant tactile images. There are two main families of such camera-based tactile sensors: marker-based, represented by TacTip sensors~\citet{TacTipFamily}, and image-based, represented by GelSight sensors~\citet{gelsightFamily}. The former tracks markers printed on the inner side of the sensor's membrane, while the latter considers the entire raw image for photometric analysis. Both of the two families started from the design of having the membrane placed on a flat clear substrate. However, in recent years a few variants, with varied morphologies, have been proposed to accommodate the needs of different applications~\cite{gomes2020geltip,omnitact,softRoundGelSight}.

Similar to many other tactile sensors that use a soft membrane to conform to sensed objects, camera-based tactile sensors suffer from degeneration through use due to the wear-and-tear of the membrane. To mitigate this issue, one solution is to run experiments with such sensors in simulation first, and then deploy the trained models for the final evaluation with the real sensors. Compared to running the entire work using the real setup, the Sim2Real approach is less damage-prone to the tactile sensors, and potentially less time-consuming by the usage of parallel training in simulation. 
%When learning methods are involved, this is commonly referred to as \textit{Sim2Real} transfer learning, in which the robotic agent is initially trained in simulation and only fine-tuned or evaluated with the real hardware.  
%  and we are particularly interested in the GelSight sensors that can capture high resolution tactile images. 
% A simulation model was first introduced in~ and extended in~\citet{}, for simulating flat GelSight sensors.
To this end, in recent years simulation models have been proposed for camera-based tactile sensors, particularly GelSight sensors \citet{gomesgelsight, gomesGelSightRAL, phisicsBasedLight, tacto,chen2023tacchi}, which have been widely used~\citet{seeThroughSkin, zhao2023skill,jiang2022shall} and are of our interest.

%To this end, in the recent years simulation models have been proposed for camera-based tactile sensors and we are particularly interested in the GelSight sensors that can capture high resolution tactile images. We introduced a simulation model for the first time in~\citet{gomesgelsight} and extended in~\citet{gomesGelSightRAL}, for simulating flat GelSight sensors. It has been widely used to simulate GelSight-like sensors~\citet{seeThroughSkin,belousov2021architectural}.

In this paper, we propose a novel geodesic-based approach to simulate camera-based tactile sensors of complex morphologies, particularly ones in which the light travels in the membrane placed on a curved surface. We consider the geodesic, i.e., the shortest path on a curved surface, as a reasonable path for the light to travel in the membrane. Since the light within the entire sensor is constant, we pre-compute the light field that is generated from each light-source beforehand, i.e., a process that is commonly referred to in computer graphics as \textit{light baking}. In this way, we only need to compute the geodesic once and we can generalise the previously proposed GelSight simulation method~\citet{gomesgelsight, gomesGelSightRAL} from a constant to a non-linear light field, enabling to simulate curved sensors such as~\citet{gomes2020geltip, softRoundGelSight}. Furthermore, as we make use of geodesic paths for computing the light field from a mesh description of the sensor, this method can be used for simulating other existing optical tactile sensors of a complex morphology or experimenting new morphology designs in simulation before fabrication of the sensor.

To evaluate our proposed method, we collect a dataset of real RGB tactile images using an improved GelTip sensor~\citet{gomes2020geltip}, and corresponding synthetic depth-maps that are then used to generate synthetic RGB tactile images. As shown in Figure~\ref{fig:cover}, a 3D printer was augmented with two additional servo motors to perform accurate tapping motions throughout the entire sensor surface, and the corresponding setup is implemented in the widely used simulator MuJoCo \citet{mujoco}. Both qualitative and quantitative analyses are performed to compare the difference between real and virtual tactile images, which is as low as \textbf{3.9\%} on average in the Mean Absolute Error (MAE) and a similarity of \textbf{0.93} in the Structural Similarity Index Measure (SSIM). Furthermore, to demonstrate the capability of this method being applied for Sim2Real learning, a contact localisation task is carried out: The model is trained with tactile images generated by our simulation method, and is then evaluated using the real samples from the GelTip sensor.

\begin{figure}[t]
\centering
\includegraphics[width=0.49\textwidth]{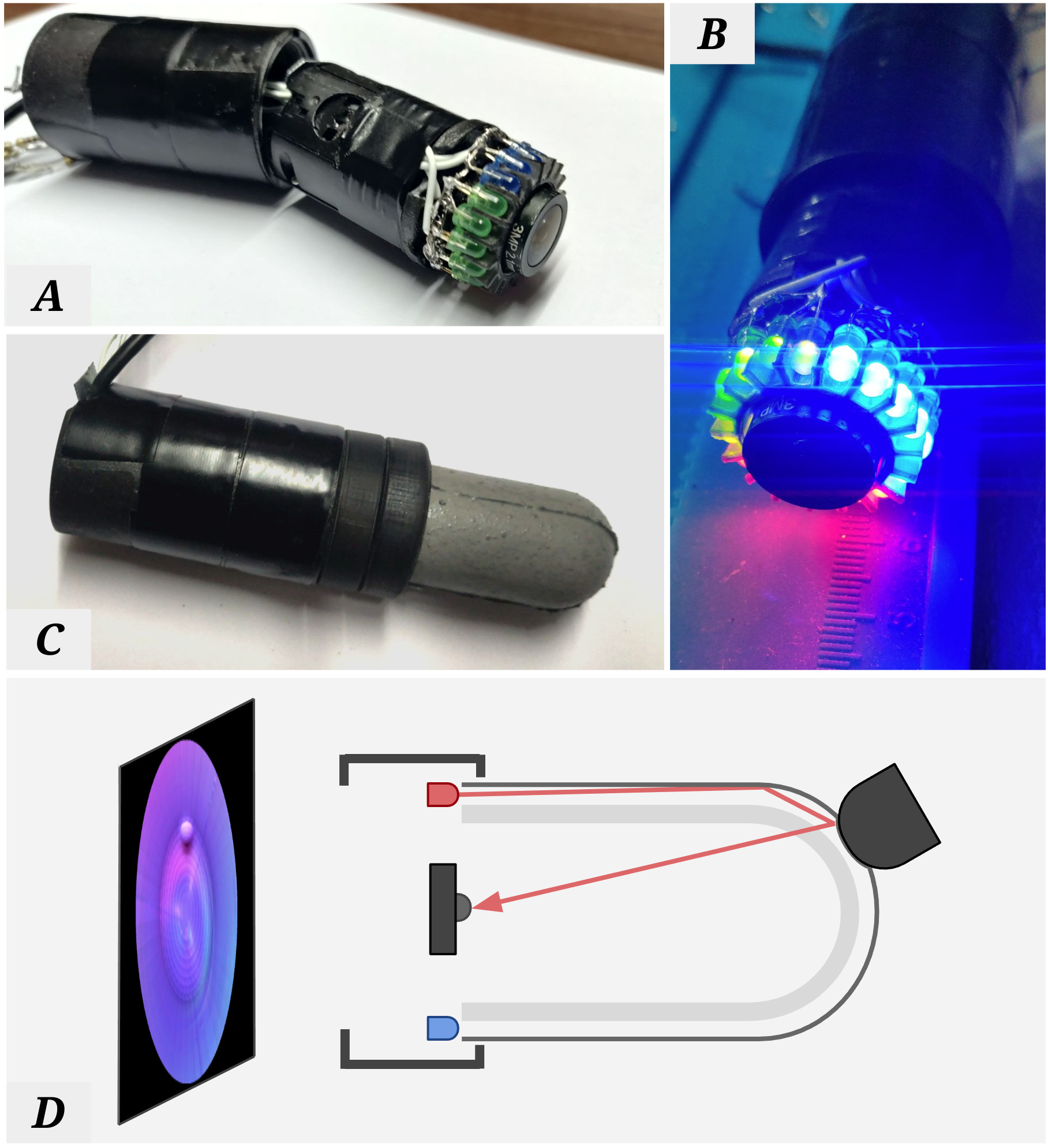}
\caption{The GelTip sensor used to capture the real tactile images. The used sensor results from a few modifications over the previously proposed GelTip \citet{gomes2020geltip} sensor that aim at improving its practical robustness. Specifically, the original camera mount is used to secure the camera boards (A); the body of the sensor is simplified into a single cylindrical case (A, C); and the positioning and quantity (from 3 to 6, per colour) of the LEDs (A, B) are revised for improved illumination. (D) The sensor working principle: light rays emitted by the LEDs travel through the elastomer, around the finger-shaped membrane. An object contacting the elastomer membrane distorts the elastomer, affecting the light path and resulting in the tactile imprint captured by the sensor camera.} 
\label{fig:used_sensor}
\end{figure}

\section{Related works}

\subsection{High-resolution optical tactile sensors}

The \textit{GelSight} working principle has been initially proposed in \citet{RetrographicSensing} as a method for reconstructing the texture and shape of contacted objects. To that purpose, three light sources are placed from opposite angles next to a transparent elastomer that is coated with an opaque reflective paint, resulting in three different shaded images of the in-contact object texture. Because of the constrained setup, a direct mapping between the observed image pixel intensities and the elastomer surface orientation can be established, enabling the surface to be reconstructed using photometric analysis. Thanks to its simple fabrication and the high-resolution of tactile images for robots to extract rich information, a wide variety of designs have been proposed to use this working principle to construct robotic tactile sensors~\citet{gelsight2017,digit,GelSlim,gelslim3}. Most of these designs focused on miniaturising the sensor and they have been extensively investigated for various applications~\citet{stamTextureRecognition, jiang2021vision, lee2019touching, jiang2022shall, jing2023unsupervised, zhao2023skill, graspRegrasp,cao2023vis2hap}. However, most of them are constrained with a single flat sensing area, which limits the potential of these sensors being used in unstructured environments where unexpected contacts can occur either inside or outside of the grasp closure~\citet{BlocksWorldOfTouch}.
To this end, a few designs have been recently proposed, in which a highly curved and/or domed finger-shaped surface membrane is considered instead, e.g., the GelTip sensor~\citet{gomes2020geltip,BlocksWorldOfTouch}, the OmniTact sensor~\citet{omnitact} and a semi-round sensor~\citet{softRoundGelSight}. GelTip and \citet{softRoundGelSight} follow the original design of the GelSight sensor \citet{gelsight2017}, using glass spectres to guide the light through the membrane. In contrast, the OmniTact points the lights directly onto the membrane surface, which was also applied in flat sensors such as~\citet{GelSlim, digit}. However, as argued and validated in~\citet{wang2021wedge}, it is highly desirable to have the light travel through the membrane and tangent to the sensing surface, so as to achieve a more homogeneous light distribution and enable using methods such as Poisson surface reconstruction for reconstructing the elastomer surface.

\subsection{Simulation of GelSight sensors for Sim2Real learning}

To save time and resources, it is desirable to develop and test robotic agents initially within a simulator before their deployment in real life. However, existing simulators lack appropriate models to simulate optical tactile sensors. To address this challenge, an approach was proposed to use a depth sensor in simulation to capture the geometry of the in-contact object and then render the obtained depth map to get a realistic tactile image for simulating GelSight sensors in~\citet{gomesgelsight,gomesGelSightRAL,chen2023tacchi}. 
Some other works used OpenGL directly~\citet{tacto} or physics based light modelling~\citet{phisicsBasedLight} to simulate the GelSight sensors. However, the latter methods are constrained to simulating sensors in which the light is shone directly at the sensor's membrane~\citet{gelsight2017, digit, omnitact}, not through the membrane~\citet{gelsight2017, softRoundGelSight, wang2021wedge}. It is vital to have the light travel through the membrane, to ensure a homogeneous distribution of light on the membrane for better surface reconstructions, as argued in~\citet{wang2021wedge}. This has particular implications for curved sensors such as~\citet{gomes2020geltip, softRoundGelSight}, as computing all the linear reflections within the curved membrane is computationally infeasible, and thus it is necessary to study the curved path through which the light travels within the membrane.

\begin{figure}
\centering
\includegraphics[width=0.48\textwidth]{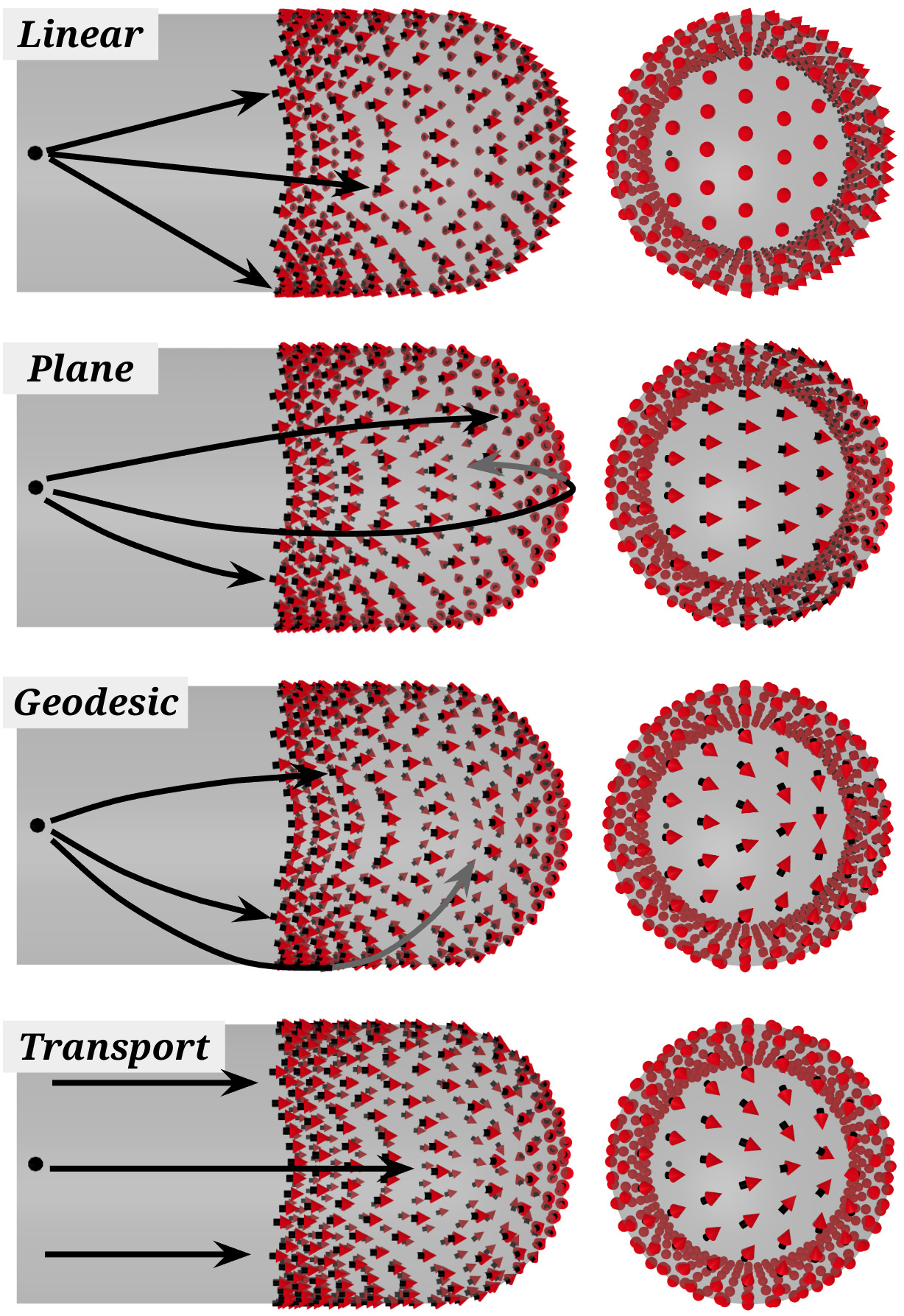}
\caption{The four light fields $\hat{L}_m$, for a given light source represented by the black sphere. The \textbf{linear} light field is highly non-tangent to the surface of the sensor around the tip, which results in the bright illumination shown in Figure~\ref{fig:samples}. In contrast, the remaining \textbf{Plane}, \textbf{Geodesic} and \textbf{Transport} fields are always tangent to the surface of the sensor, resulting in no illumination in areas without deformation caused by contacts.}
\label{fig:vector_fields}
\end{figure}

\section{Method}
\label{sec:method}

Starting from the method proposed in \citet{gomesgelsight, gomesGelSightRAL} for the flat GelSight sensors, additional steps are needed to address the curved geometry of the sensor and the light being guided through the curved membrane when simulating optical tactile sensors whose surface is not flat, for example the GelTip sensor as illustrated in Figure~\ref{fig:used_sensor}. In these sensors, the orthographic camera assumption is dropped, and the camera is explicitly modelled using the standard camera model. More importantly, the light directions are generalised from constants to light fields. We compute the light fields using 4 different methods that we will refer throughout the paper as \textit{Linear}, \textit{Plane}, \textit{Geodesic} and \textit{Transport}. The overall approach of simulating optical tactile sensors consists of two steps: the light fields of the sensor are first computed offline; with the computed light fields, the simulation model is then run online for each frame of the sensor simulation. The online simulation model consists of three main steps: 1) smoothing of the raw depth map, captured from the simulation environment, to mimic the elastic deformation of the real sensor; 2) mapping of the smoothed depth map onto a point-cloud in the camera coordinates frame; 3) generation of the tactile image using Phong's illumination model and the pre-computed light fields.

\subsection{Offline pre-computation of light fields}
% http://www.cs.umd.edu/~mount/Papers/mmp-sicomp-87.pdf
\label{subsec:method_light_field}
A light ray emitted by a point light source $L_m$ to a target point $T$ in open space travels in a straight direction, \mbox{$\vv{L} = L_m - T$}, resulting in a \textit{Linear} light field of radial or conical shape, or a uniform light field (of parallel rays) if $L_m$ is considered at the infinity. However, for some GelSight sensors such as \cite{gomes2020geltip, softRoundGelSight, gelsight2017}, because the illumination sources are placed tangential to the tactile membrane, the light field travels through the membrane and assumes the sensor surface geometry. For curved sensors \cite{gomes2020geltip, softRoundGelSight}, this results in a highly non-linear light field $\hat L_m(u,v)$ that must be computed taking into consideration the sensor's tactile membrane geometry:
\begin{align}
\label{light_field_eq}
\hat L_m(u,v) = f(L_m, P, M)
\end{align}
% where $L_m$ is the light source position, $P$ is the point-cloud derived from \shan{the depth map $D_{deform}(u,v)$} and $M$ is the sensor surface mesh description.
% We assume that on a curved surface, the predominant ray traveling from $L_m$ to $P$, through the transparent elastomer, follows the corresponding shortest path possible, or geodesic.}
where $P$ is the point cloud derived from the depth map $D_{deform}$ (\cite{gomesGelSightRAL} and detailed in Sub-section~\ref{sec:deformation}) and $M$ is the sensor surface's mesh description.

We assume that on a curved surface, the predominant ray travelling from the point light source $L_m$ to the target point $T$, through the transparent elastomer, follows the corresponding shortest path possible, or geodesic that is a locally length-minimising curve. To compute such light paths from the sensor membrane mesh description, one naive approach is to intersect the mesh of the tactile membrane with a \textit{Plane} that contains the light source position $L_m$ and the corresponding target $T$ and is orthogonal to the mesh surface on $L_m$. The final sub-segment between $L_m$ and $T$ can then be derived and sorted, followed by extracting the final $\hat L_m$. Figure~\ref{fig:vector_fields}-A shows the resulting light fields for a single light source over the mesh using this \textit{Plane} method. 

A more accurate discrete \textit{Geodesic} path can also be computed, using the method proposed in \citet{Mitchell1987TheDG}. The algorithm works similarly to the Dijkstra shortest path algorithm for graphs, in which the distance between two neighbouring vertices is incrementally propagated through the graph. Figure~\ref{fig:vector_fields}-B shows the resulting light fields for a single light source over the mesh using this \textit{Geodesic} method. However, the computation of such geodesic paths is computationally expensive and while the light travels effectively through such geodesic paths, for the illumination rendering purposes we are only concerned with the vector at the end of such paths. In other words, we are concerned with the problem of vector \textit{Transport}. In \citet{Sharp:2019:VHM}, the vector heat method was proposed, in which these vectors can be computed more efficiently via short-time heat flow, resulting in a significantly more efficient algorithm.

While the \textit{Transport} method is highly more efficient than the \textit{Planes} and \textit{Geodesic} methods, reducing the computation time from around one and a few hours respectively, to a few minutes (considering our working parameters, mesh, depth-map resolution, parallelisation over an 8-core CPU, etc.), it is still infeasible to be run in real time for processing the light fields of each frame of the tactile images. Nonetheless, as GelSight sensors are enclosed and immune to external light variance, and that the elastomer membrane sits on top of a rigid glass, the deformations of the membrane during contacts would be insufficient to alter the overall light field, and as such we can assume that the light field stays constant. To this end, we can compute the entire light field only once, offline, and store the resulting light map for later online use. %Figure~\ref{fig:vector_fields} shows the resulting light fields for a single light source over the mesh, and these are further analysed in the experiments section.

\begin{figure}
\centering
\includegraphics[width=0.48\textwidth]{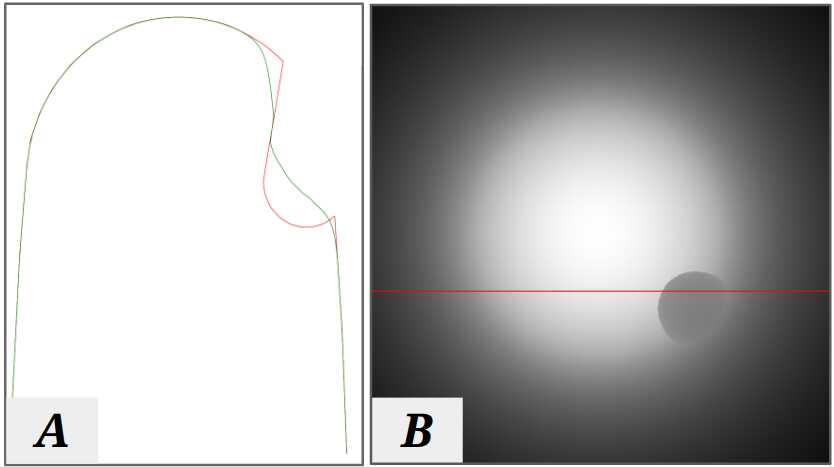}
\caption{(A) The profile of the raw (in red) and Gaussian smoothed (green) depth maps, for the row highlighted in red in~(B). As shown in the figure, the green smoothed profile resembles the real deformation more closely. (B) The raw depth map captured from the depth camera installed inside the sensor membrane in the MuJoCo environment, whilst a sphere protrudes the sensor membrane. }
\label{fig:elastic_deformation}
\end{figure}

\subsection{Elastomer elastic deformation approximation}
\label{sec:deformation}

While running online, a depth-map $D$ is captured from the simulation environment and its elastic deformation $D_{deform}$ is approximated with a Gaussian filter, followed by an element-wise minimum operation to ensure the in-contact protrusion stays undeformed, similarly to  \citet{gomesgelsight, gomesGelSightRAL}: 
\begin{align}
% D_{protrusion} &= D_{ref} - D\\
% % G(x,y) &= \frac{1}{\sqrt{2\pi\sigma^2}}  e^{-\frac{x^2+y^2}{2\sigma^2}} \\
% D_{deformation} &= D_{protrusion} \ast G    \\
% D_{deformed} &= min(D, D_{ref} + D_{deformation}) 
D_{deform} &= min(D, D_{ref} + (D_{ref} - D) \ast G)
\label{eq:filter}
\end{align} 
% take out the \ in Dref when removing the revised.
% In which, $D$ is the depth map captured online within the simulation, $D_{ref}$ is a reference depth map taken without any contact occurring and $G$ is the Gaussian kernel convolved over the previous depth maps difference. 
\noindent where $D$ is the depth map captured online within the simulation, $D_{ref}$ is a reference depth map taken without any contact occurring and $G$ is the Gaussian kernel convolved over the difference of $D$ and $D_{ref}$. Figure~\ref{fig:elastic_deformation} shows an example of the result of such operations over a profile of the depth map being contacted by a sphere object.

\subsection{Inverse camera projection model}
The elastomer surface height map captured in the form of a depth-map \mbox{$D_{deform}$} is then converted into a point cloud \mbox{$P(u, v)$} to enable the later application of the Phong's Illumination model \citet{phongIlluminationModel}: % and ensure the proper alignment with the mesh used in 

% ref : http://www.open3d.org/docs/release/python_api/open3d.geometry.PointCloud.html#open3d.geometry.PointCloud.create_from_depth_image
\begin{align}
P(u, v) = \systeme*{
% / depth_scale
x = (u - c_x) \frac{z}{f_u},
y = (v - c_y) \frac{z}{f_v},
z = D_{deform}{(u, v)}
}
\label{eq:proj_ray}
\end{align}
where $(c_x, c_y)$ is the centre of the depth map $D_{deform}$; $f_u$ and $f_v$ are given by:

\begin{align}
\label{focal_length}
    % foradiansv = math.(fov\_deg)
f_u = \frac{D_{width}}{2 \tan(\frac{fov}{2})}, 
f_v = \frac{D_{height}}{2 \tan(\frac{fov}{2})} 
\end{align}

\noindent where $D_{width}$ and $D_{height}$ are the width and height of the depth map $D_{deform}$ in pixels; $fov$ is the angular camera field of view in radians.

\subsection{Image Rendering using the Phong's reflection model}

The generation of RGB tactile images $I$ from the height-map $H$ of the elastomer (obtained from \mbox{$D_{deform}$}) can be interpreted as the inverse problem of the surface reconstruction problem~\citet{RetrographicSensing}, as the former consists of finding the mapping function \mbox{$H \rightarrow I$} while the latter \mbox{$I \rightarrow H$}. In both cases, the relationship between the two can be described by:
\begin{equation}
\label{eq:eq_1}
I(x,y) =  R\left(\frac{\partial H}{\partial x}, \frac{\partial H}{\partial y}\right)
\end{equation}
where $R(\cdot)$ is the reflectance function that models both the lighting conditions (i.e., illumination of the LEDs) and the reflectance properties of the surface material (i.e., the elastomer coating paint). Here, it should be noted that the colour observed at a given pixel is directly correlated with the orientation of the corresponding point on the elastomer.  
In~\citet{RetrographicSensing}, the mapping of the two points in the image space and the elastomer is built through a calibration process. In our case, we get $R(\cdot)$ using the Phong's illumination model~\citet{phongIlluminationModel}. Phong's model is an empirical model of local illumination that has been developed in the context of 3D Computer Graphics to describe how a given surface reflects light as a combination of the diffuse and specular reflections. 
From Phong's model, $I(x,y)$ observed at a given point of the sensor elastomer is given by three components: ambient, diffuse and specular light, as
\begin{equation}
I =k_a i_a+\sum_{m \in L}{(k_d(\hat L_m \cdot \hat N) i_{m,d} + 
k_s(\hat R_m \cdot \hat V)^\alpha i_{m,s})} \label{eq:phong_model}
\end{equation}
\begin{equation}   
\hat R_m = 2 (\hat L_m \cdot  \hat N ) \hat N - \hat L_m
\end{equation}
where $L$ is the set of light sources (i.e., LEDs), $\hat L_m$ is the emission direction of a given light source $L_m$; $i_a$ is the intensity of the ambient light that is not caused by any of the LEDs; $i_{m,d}$ and $i_{m,s}$ are the intensities of the diffuse and specular reflections of light source $m$ respectively; $k_a$, $k_d$, $k_s$ and $\alpha$ are all reflectance properties of the surface; $\hat R_m$ is the direction that a perfectly reflected ray of the light would take; $\hat V$ is the direction pointing towards the camera. Given that our camera is pointing perpendicularly to the elastomer, $\hat V$ is set to $<0,0,1>$. The normalised surface normals $\hat N$ are computed using the discrete partial derivatives of the height-map, as in the surface reconstruction~\citet{RetrographicSensing}:
\begin{align} 
\hat N &= \;\; \frac{\frac{\partial p}{\partial x} \times \frac{\partial p}{\partial y}}{
\|\frac{\partial P}{\partial x} \times \frac{\partial p}{\partial y}\|
} \\
% &= \;\; < \frac{H}{2r} \ast \begin{bmatrix}-1 &\!\!\!\! 0 &\!\!\!\! 1\end{bmatrix} , \frac{H}{2r} \ast  \begin{bmatrix}-1 &\!\!\!\! 0 &\!\!\!\! 1\end{bmatrix}^T, -1 >
\nonumber
\end{align} 
where the partial derivatives $\frac{\partial p}{\partial x}$ and $\frac{\partial p}{\partial y}$ are computed using the Sobel edge detector over a point $p$ in the point cloud $P$. 
Given the closed design of optical tactile sensors, and to capture unmodeled  illumination, we set the background illumination $k_ai_a$ to be an aligned image captured using a real sensor, when no contact is being applied to it.

%\begin{figure}
%\centering
%\includegraphics[width=0.48\textwidth]%{figures/vector_fields.jpg}
%\caption{(A) The \textbf{Linear} and (B) %\textbf{Geodesic} light fields $\hat{L}_m$, for a %given light source represented by the black sphere. %As shown, the linear light field is highly non %tangent to the surface of the sensor, around the %tip, which results in the bright illumination shown %in Figure~\ref{fig:samples}, whilst in contrast, %the \textbf{Geodesic} field, is always tangent to %the surface, resulting in no illumination in areas %without deformation caused by contacts. (C) The %surface normals, $\hat{N}$, computed online from %the discrete partial derivatives of the deformed %sensor surface $P$.}
%\label{fig:vector_fields}
%\end{figure}

\begin{figure}
\centering
\includegraphics[width=0.4\textwidth]{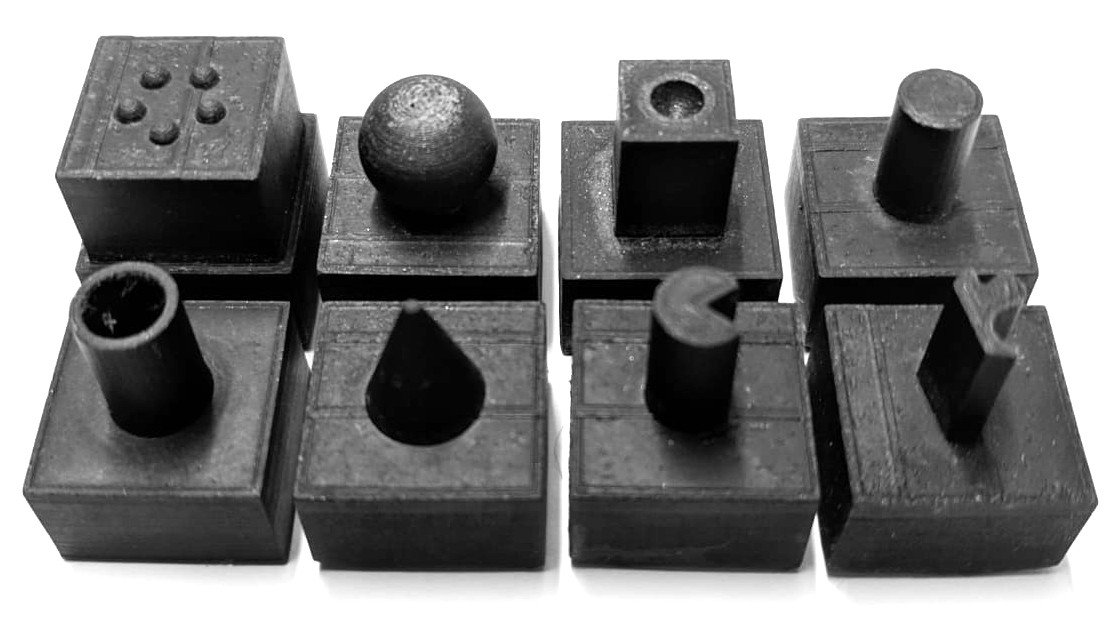}
\caption{Set of 3D printed objects, used to collect the  dataset of tactile images. From left to right, top to bottom, ``dots'', ``sphere'',  ``indented prism'', ``cylinder'', ``open cylinder'', ``cone'', ``pacman'' and ``random''.}
\label{fig:object_set}
\end{figure}

%\begin{table}
%\centering
%\caption{Configurations for the illumination}
%\def\arraystretch{1.2}
%\begin{tabular}{  l | l  c  c  c } 
% & $i_{m,s}$, $i_{m,d}$ \tiny{(RGB)} & $k_d$ & $k_s$  \\
%\hline
%Blue & $(87, 159, 233)$ & $0.95$ & $0.1$ \\
%Green & $(197, 226, 241)$ & $0.7$ & $0.1$ \\ 
%Red & $(196, 94, 255)$ & $0.95$ & $0.1$ \\ \hline
%Solid bg. & $i_a=(153, 128, 245)$ & \multicolumn{2}{c}{$k_a=0.8$}\\
%\end{tabular}
%\label{table:lights_configs}
%\end{table}

\section{Experimental Setup and GelTip Dataset}

To evaluate our simulation method, we collect aligned datasets of contacts on the real and simulated GelTip sensors, by tapping 8 3D printed objects used in~\citet{gomesGelSightRAL}, as shown in Figure~\ref{fig:object_set}, against the sensor.
The data collection motions consist of equidistant and perpendicular taps on 3 straight paths along the longer axis of the sensor, from the tip to its base, with the sensor rotating $\pi/2$ in between each path.

\begin{figure*}[t]
\centering
\includegraphics[width=\textwidth]{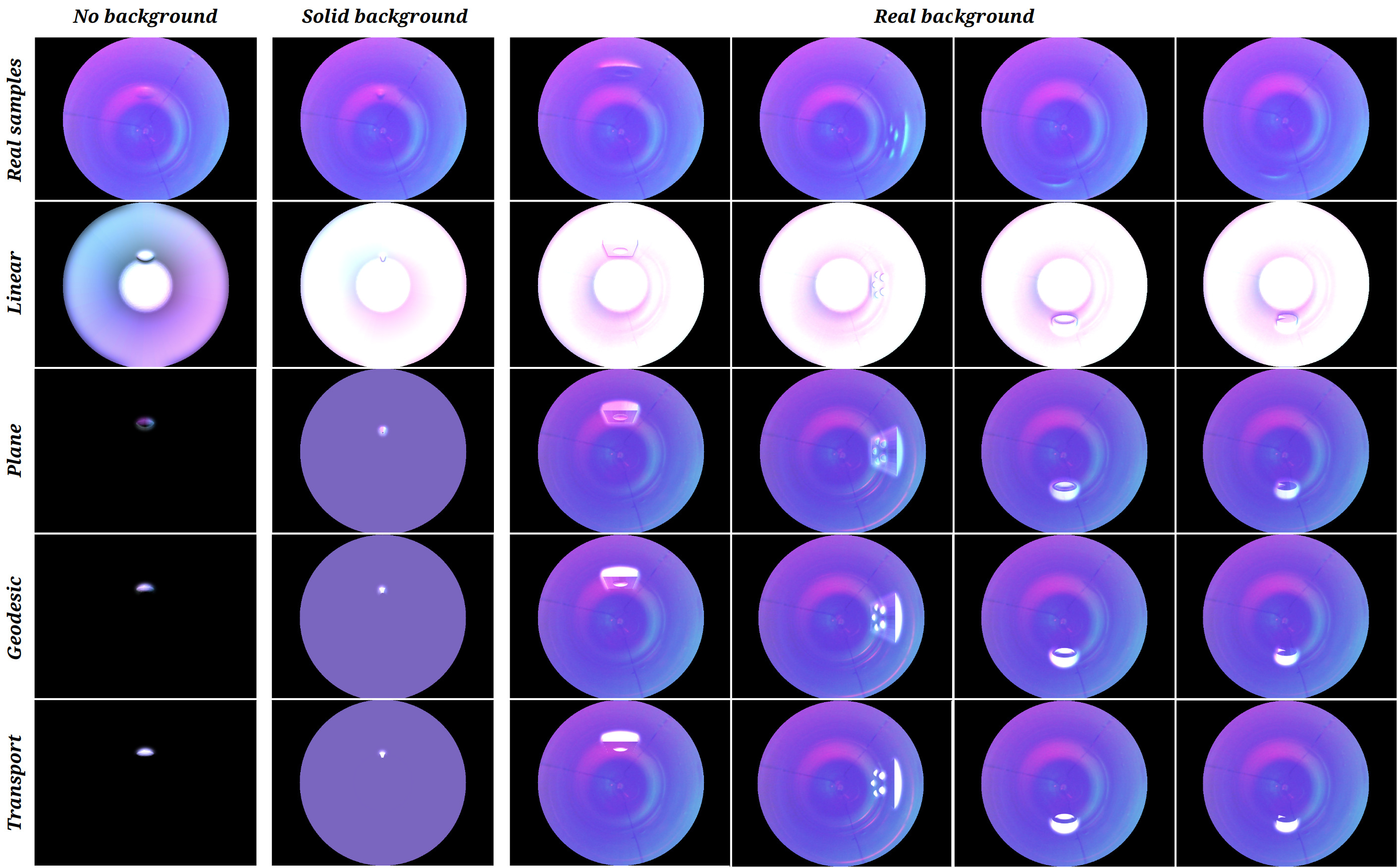}
\caption{Samples of tactile images extracted from the real (aligned) and synthetic datasets. From left to right, contacts against the ``sphere'',  ``cone'', ``indented prism'', ``dots'' and ``open cylinder'', from the object set shown in Figure~\ref{fig:object_set}. From top to bottom: the real samples; the simulations using the \textit{Linear},  \textit{Geodesic} and \textit{Transport} light fields. As highlighted in the first column, with no ambient illumination applied, the \textit{Geodesic}, \textit{Linear} and \textit{Transport} fields result in illumination only in the in-contact areas and thus can successfully be combined with the real background image, to produce a realistic-looking simulated image, as shown in the three columns on the right. In contrast to the results from the other three methods, the \textit{Linear} light field results in a highly bright tip, which is different from what occurs in real images. 
}
\label{fig:samples}
\end{figure*}

%\subsection{Experimental Setup}
In the real environment, the captured tactile images are the RGB images from the sensor camera, while in simulation, are from the depth maps in MuJoCo. Then, we experiment with the simulation model offline by setting its hyper-parameters, e.g., the different light fields, and generate the corresponding RGB synthetic tactile images. To capture the real samples, we set up a \textit{GelTip} sensor on a Fused Deposition Modelling (FDM) 3D printer (Anet A30, model from Geeetech) and fix the contacting object to the printer head.
Due to the curvature of the \textit{GelTip} and the necessity of generating curved and downward motions normal to the sensor surface, two additional servos Hitec HS-422 are used to control the orientation of the tactile sensor ($S_{g}$) and of the contacting object ($S_{i}$). The printer is controlled by issuing \textit{G-code} commands through serial communication. Similarly, the servos are controlled using an Arduino Mega 2560 board. When setting up the simulation in MuJoCo, the parameters to take into consideration are the visual map $z_{near}$ and $z_{far}$ attributes that set the minimum and maximum depths of the OpenGL projection clipping planes, as well as the camera's Field of View (FoV). $z_{near}$ plane is carefully adjusted to ensure that the \textit{GelTip} membrane is entirely rendered. Further, $z_{near}$ and $z_{far}$ values must be later used to un-normalise the initially normalised depth map $D_{[0,1]}$ provided by the MuJoCo API following:
\begin{align} 
D = \frac{z_{near}}{1 - D_{[0,1]} \frac{1 - z_{near}}{z_{far}}}
\end{align} 

The simulation depth maps and real tactile images, are manually aligned by visually finding the transform that produces highest overlapping between real and simulated contacts.

To compute the light fields, a reference depth map (with no contacts) is captured from the simulator and converted into a point cloud. This point cloud is then used to align the sensor elastomer mesh (that was used to construct the real and simulated sensors). Finally, the light fields are computed. The \textit{Linear} light field is computed straightforward as \mbox{$\hat{L}_m = \frac{L - p}{\|L-p\|} $}, where $p$ is a point on the sensor surface and $L$ is the position of the light source. The \textit{Plane} field is computed using the Python library \textit{trimesh}\footnote{https://pypi.org/project/trimesh/} for solving the plane-mesh intersection, followed by a heuristic algorithm to extract the vector at the target point, as described in Section~\ref{subsec:method_light_field}. The \textit{Geodesic} field is computed using \textit{pygeodesic}\footnote{https://pypi.org/project/pygeodesic/} and the \textit{Transport} using the transport vector method from \textit{potpourri3d}\footnote{https://pypi.org/project/potpourri3d/}.
These real and simulated datasets, together with object STL files, and source-code can be found on our project website.

\begin{table*}[t]
\centering
\caption{Real and generated datasets comparison}
\def\arraystretch{1.3}
\begin{tabular}{ l | r r r | r r r  | r r r } 
 & 
 MAE & SSIM & PSNR &
 MAE & SSIM & PSNR &
 MAE & SSIM & PSNR \\ \hline
% \bm for bold. 
% \pm for +- e.g. ${\bm{8.40\pm0.04}}$
& 
\multicolumn{3}{c |}{\textsc{no background}} &
\multicolumn{3}{c|}{\textsc{solid background}} &
\multicolumn{3}{c}{\textsc{real background}} \\ \hline

Linear   & 
\bm{$10.9$\%} & \bm{$0.84$} & \bm{$14.83$} &
$17.6$\% & $0.85$ & $10.54$ &
$17.4$\% & $0.85$ & $10.90$ \\

Plane     & 
$37.5$\% & $0.39$ & $5.50$ &
\bm{$7.5$}\% & \bm{$0.93$} & \bm{$18.28$} &
\bm{$3.9$}\% & \bm{$0.94$} & \bm{$24.86$} \\

Geodesic  & 
$37.5$\% & $0.34$ & $5.50$  &
$7.6$\% & \bm{$0.93$} & $18.13$ &
\bm{$3.9$}\% & $0.93$ & $24.20$ \\

Transport & 
$37.6$\% & $0.38$ & $5.49$ &
$7.6\%$ & \bm{$0.93$} & $18.10$ &
$4.0\%$ & \bm{$0.94$} & $24.08$ \\

\end{tabular}
\label{table:fields_losses}
\end{table*}

\section{Experiments}

In this section we evaluate our method by firstly analysing and comparing the synthetic datasets generated using the different light fields, secondly demonstrating the usage of the synthetic dataset to pre-train a neural network in Sim2Real tasks and, finally, illustrating the application of the proposed simulation model to explore possible future sensor designs.

\subsection{Quantitative and qualitative comparison of the light fields}
To assess the four different light fields, we generate the corresponding version of the dataset, from the captured depth maps in simulation, and quantify how similar the generated tactile images are, when compared against the real correspondences, by computing the Mean Absolute Error (MAE), Structural Similarity (SSIM)~\citet{ssim} and Peak Signal-to-Noise Ratio (PSNR) over the entire dataset for the different scenarios. 
As reported in Table~\ref{table:fields_losses}, the \textit{Linear} light field results in better MAE and SSIM of $10.9$\% and $0.84$, when no ambient light is considered. However, if either a solid ambient illumination, or the background image from the real sensor is used, then the \textit{Plane}, \textit{Geodesic} and \textit{Transport} methods produce better results, with a small MAE of $3.9$\% and a high SSIM of $0.93$. This directly comes from the fact that the \textit{Plane}, \textit{Geodesic} and \textit{Transport}
methods do not produce any illumination in areas that are not being deformed by a contact, which contrasts with the \textit{Linear} method that generates bright gradients throughout the entire sensor surface. This occurs because when no ambient light is considered, $k_a i_a = 0$, and where no contacts are happening, with the light travelling tangent through the surface,  $\hat{L}_m \perp \hat N \implies \hat L_m \cdot \hat N = 0$, and the entire Phong's illumination expression is zero. This happens by design and follows directly from the GelSight working principle \citet{RetrographicSensing}. 

However, when compared against the real tactile images, we find that in practice much light travels linearly from the light sources through the membrane core, due to the sub-optimal construction of the GelTip membrane. As shown in the examples with a real background image in Figure~\ref{fig:samples}, this results in the highly bright and colourful gradients in areas without any contacts and contrasting to an ideal lighting that would only travel through the elastomer membrane. Nonetheless, the \textit{Linear} light field results in a highly bright tip of the sensor, which heavily contrasts with the more homogeneous (or even darker) tip observed in the real images.
These gradients in areas of no contact can also be observed in flat GelSight sensors, due to their imperfect construction. In the flat GelSight simulation method \citet{gomesGelSightRAL}, this issue was addressed by using a real tactile image as background illumination, which is captured when no contact is occurring. Here, because the non-linear fields are tangent to the elastomer surface, we can use the same approach, as shown in Figure~\ref{fig:samples}.

% \draft{Within the three non-linear light fields, }

%\begin{table}
%\centering
%\caption{Image classification results %summary}
%\def\arraystretch{1.3}
%\begin{tabular}{ l | r r } 
%& Train & Validation \\
%\hline
%Real2Real & \\
%Sim2Sim & \\
%\hline
%Sim2Real (patches) & \\
%Sim2Real (dataset) & \\
%\end{tabular}
%\label{table:object_localization}
%\end{table}

\subsection{Sim2Real transfer of contact perception}

To evaluate the Sim2Real transfer of models optimised using the synthetic tactile images we experiment with contact localisation and classification. To this end, the real and simulation RGB datasets are split into equal training (80\%) and validation (20\%) splits, and then a ResNet
 from \citet{resnet}\cite{torchvision2016} is trained and evaluated using either simulated or real data to investigate how the models trained in simulation will perform on the real data, compared to the performance of evaluating the performance of the trained model in its own domain. Specifically, we have three different scenarios: \textit{Real2Real} (\textbf{R2R}), \textit{Sim2Sim} (\textbf{S2S}) and \textit{Sim2Real} (\textbf{S2R}). 

 In all scenarios the network is trained with random batches of 64 images, 8 batches per epoch, for 100 epochs, using Adadelta, a learning rate of 0.1 and the mean squared error loss. The dataset computed with the \textit{Plane} light field and real background image is used for the simulation data. Further, to fully exploit the advantage of simulation, and given the small \textit{Base} real/sim datasets of $18\times8=144$ images, an \textit{Extended} simulated dataset is also collected, using the same simulation model configuration, in which the number of rows is increased to 5, the contacts per row to 16, and apply small transformations to the sensor and indenter are applied, resulting in a dataset of $2,880\times8=23,040$ synthetic images. For the classification task, we also experiment with cropping centred patches around the in-contact areas. For the localisation task, the errors reported in millimetres are computed using the camera model to project the predictions of the network in the image space to points in the camera coordinates, as in~\citet{gomes2020geltip}.
 
As reported in Tables \ref{table:object_classification} and \ref{table:object_localization}, in general, performing these tasks with the simulated images, \textit{Sim2Sim}, results in significantly better results than the corresponding \textit{Real2Real} cases. With the model achieving the overall highest accuracy of $100\%$ in the classification task, and the lowest contact localisation error of \SI{3.7}{\mm}. In contrast, in the \textit{Real2Real} scenario the model achieves only $64.3\%$ accuracy and a localisation error of \SI{8.4}{\mm}. One possible reason could be that with the real sensor, the contact forces between the in-contact object and the sensor, results in some bending of the sensor membrane and consecutive weaker imprints, as noticeable in the examples shown in Figure~\ref{fig:samples}. On the other hand, while the \textit{Extended} dataset and cropping of centred patches positively contribute to a reduction in the \textit{Sim2Real} gap, the \textit{Sim2Real} gap is still significant: with the model trained with the simulated data achieving only a maximum classification accuracy of $50.0\%$, in the classification task, versus the $64.3\%$ obtained by the model trained with the real data. In the \textit{Sim2Real} contact localisation, the model obtains a minimum localisation error of \SI{11.8}{\mm}, worse than the \SI{8.4}{\mm} obtained the model trained with real data.
%the \textit{Sim2Real} gap is significant, as the models trained with the simulated data perform significantly worse with the real data. 
%The \textit{Extended} dataset contributes to a positive reduction in the Train-Validation gap, however, it is insufficient to improve the \textit{Sim2Real} gap. 
%Similarly, the centred patches around the in-contact area further improve the \textit{Sim2Sim} results, but in opposition increase the \textit{Sim2Real} gap. 
Future research could address this \textit{Sim2Real} gap, as in ~\citet{jianu2022reducing}.

\begin{table}
\centering
\caption{Sim2Real classification Accuracy (\%)}
\def\arraystretch{1.3}
\begin{tabular}{ l |r r| r r}
& Base & Extended & Base & Extended \\
\hline
&
\multicolumn{2}{c|}{\textsc{Full Image}} &
\multicolumn{2}{c}{\textsc{Cropped Patch}}  \\
\hline
R2R  & $64.3$ &  -       & $35.7$ & - \\
S2S  & $75.0$ & $100.0$ & $96.4$ & $100.0$ \\
S2R & $42.8$  & $50.0$  & $14.3$ & $32.14$ \\
\end{tabular}
\label{table:object_classification}
\end{table}

\begin{table}
\centering
\caption{Sim2Real localisation (mm)}
\def\arraystretch{1.3}
\begin{tabular}{ l |r r}
& Base & Extended \\
\hline
R2R  & $8.4$ & - \\
S2S  & $3.7$ & $5.4$ \\
S2R  & $11.8$ &  $13.3$ \\
\end{tabular}
\label{table:object_localization}
\end{table}

%\begin{table}
%\centering
%\caption{Sim2Real results summary}
%\def\arraystretch{1.3}
%\begin{tabular}{ l |r r| r r}
%& Base & Extended & Base & Extended \\
%& \small{train} / \small{val}
%& \small{train} / \small{val}
%& \small{train} / \small{val}
%& \small{train} / \small{val} \\
%\hline
%& \multicolumn{2}{c |}{\textsc{classification Accuracy (\%)}} & \multicolumn{2}{c}{\textsc{localisation (mm)}}\\
%\hline
%R2R  & 
%$80.2$ / $64.3$ &  -       & $7.8$ / $8.4$ & - \\
%S2S  & 
%$97.4$ / $75.0$ & 
%$98.2$ / $100.0$ & 
%$3.7$ / $3.7$ & $5.2$ / $5.4$ \\
%S2R & $33.6$ / $42.8$  &   $31.6$ / $50.0$  & $9.7$ / $11.8$ &  $13.9$ / $13.3$ \\
%\hline
%& \multicolumn{4}{c}{\textsc{cropped centred patches}} \\ \hline 
%R2R & $44.8$ / $35.7$ & - \\
%S2S & $98.2$ / $96.4$ & $100.0$ /  $100.0$ \\
%S2R & 
%$20.0$ / $14.3$ & 
%$25.9$ / $32.14$ 
%\end{tabular}
%\label{table:old_object_localization}
%\end{table}

\subsection{Application to non-existing sensor designs }

\begin{figure}
\centering
\includegraphics[width=0.49\textwidth]{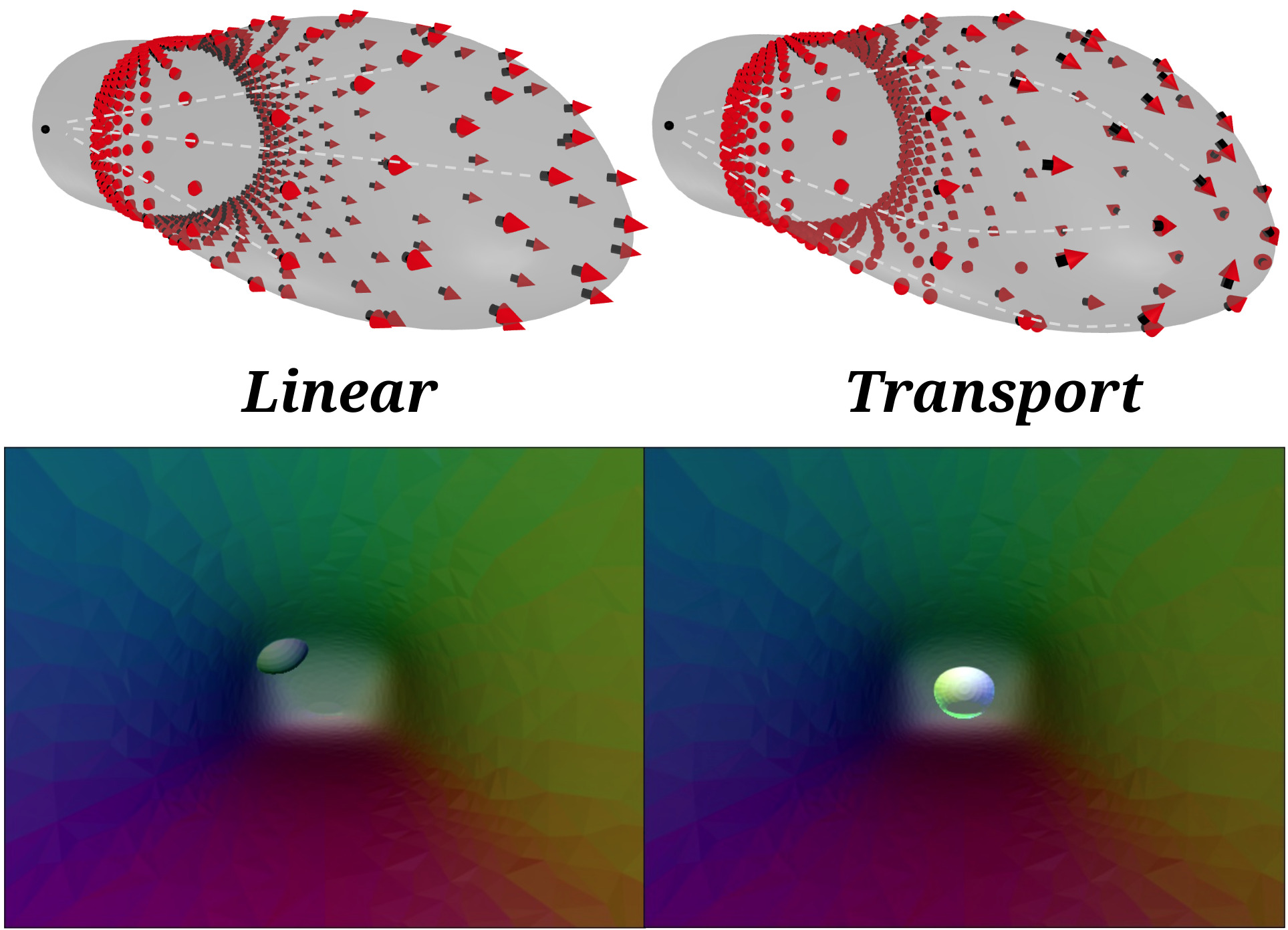}
\caption{Application of the proposed model to a non-existing sensor. Since no real sensor exists to capture the real background image, an image generated using the \textit{Linear} light field is used to create the ambient illumination for the \textit{Transport} field. }
\label{fig:thumb}
\end{figure}

With the exception of the real background image, the proposed model only depends on the sensor's mesh description and light sources parameters. Hence, it can be applied to explore future sensor morphologies, before their fabrication. To demonstrate this application, we design an irregular shaped membrane inspired by the human finger, and apply the proposed simulation model to it. As shown in Figure~\ref{fig:thumb}, the \textit{Linear} and \textit{Transport} light fields are rendered with the sensor being contacted by a sphere, and two tactile images are created in simulation using the the \textit{Linear} and \textit{Transport} light fields respectively. Since no real sensor exists to capture the real background image, an image generated using the \textit{Linear} light field is used to create the ambient illumination for the \textit{Transport} field. We can find that the \textit{Transport} method generates more realistic light paths, especially at the tip of the finger. This example shows that the simulated tactile sensor of different morphologies could be experimented to optimise its morphology for a given task. For example, for a given grasping task, the optimal morphology of the optical tactile sensor could be obtained in the simulation using our proposed simulation model.

%\begin{figure}
%\centering
%\includegraphics[width=0.49\textwidth]{figures/printers.jpg}
%\caption{Overview of the setup in simulation, with 20 printers and GelTip sensors running in parallel and faster than real time collecting the augmented dataset.}
%\label{fig:printers}
%\end{figure}

\section{Conclusion} 
\label{sec:conclusion}

In this paper, we proposed a novel approach for simulating \textit{GelSight} sensors of complex geometry, such as the \textit{GelTip}. The development of the simulation models enables to more rapidly prototype optical tactile sensors with low cost, which is an advantage for all kinds of research using optical tactile sensors. A second benefit of our proposed method is that it enables rapid design of new sensor geometries and also evolutionary designs for specific applications. Finally, the specific considerations of the light trajectories within the tactile membrane help us in better verify whether our assumptions about the real sensor are true. For instance, with the analysis of different types of light fields, we verify that tactile images captured by the existing \textit{GelTip} sensors contain a high degree of light that does not travel parallel to the sensor surface, as idealised by the early GelSight working principle \citet{RetrographicSensing}. 

From a different perspective, we can understand the tactile image illumination as resulting from multiple reflections and phenomena. By simulating the component that travels through the membrane and only affects the deformed areas, it enabled us to construct a simulation that can take advantage of the real background image, and thus result in a highly realistic simulation, as demonstrated by our quantitative results. Furthermore, the proposed method not only unlocks simulating existing optical tactile sensors of complex morphologies, but also enables experimenting with sensors of novel morphologies, before the fabrication of the real sensor. 

In the future, we will compare the tactile images obtained from a real sensor fabricated using the morphology design optimised in the simulation, and also apply our proposed simulation model in Sim2Real learning for tasks like robot grasping and manipulation with optical tactile sensing.

%\section*{Acknowledgements}

%% Use plainnat to work nicely with natbib. 

% USE \citet vs \cite 

%{\small
%\begin{verbatim}
%@article{McGeer01041990,
%  author = {McGeer, Tad}, 
%  title = {\href{http://ijr.sagepub.com/content/9/2/62.abstract}%{Passive Dynamic Walking}}, 
%  volume = {9}, 
%  number = {2}, 
%  pages = {62-82}, 
%  year = {1990}, 
%  doi = {10.1177/027836499000900206}, 
%  URL = {http://ijr.sagepub.com/content/9/2/62.abstract}, 
%  eprint = {http://ijr.sagepub.com/content/9/2/62.full.pdf+html}, 
%  journal = {The International Journal of Robotics Research}
%}
%\end{verbatim}
%}
%\noindent
%and the entry in the compiled PDF would look like:

%\def\tmplabel#1{[#1]}

%\begin{enumerate}
%\item[\tmplabel{1}] Tad McGeer. \href{http://ijr.sagepub.com/content/9/2/62.abstract}{Passive Dynamic
%Walking}. {\em The International Journal of Robotics Research}, 9(2):62--82,
%1990.
%\end{enumerate}
%
%where the title of the article is a link that takes you to the article on IJRR's website. 

%Linking cited articles will not always be possible, especially for older articles. There are also often several versions of papers online: authors are free to decide what to use as the link destination yet we strongly encourage to link to archival or publisher sites (such as IEEE Xplore or Sage Journals).  We encourage all authors to use this feature to the extent possible.

\bibliographystyle{plainnat}
\bibliography{references}

\end{document}